\documentclass{article}
\usepackage[T1]{fontenc}
\usepackage[utf8]{inputenc}
\usepackage{lmodern}
\usepackage{times}    
\usepackage{helvet}  
\usepackage{courier}  
\usepackage{graphicx} 
\usepackage{amsmath}
\usepackage{geometry}            
\usepackage{amsmath,amssymb}  
\usepackage{booktabs}   
\usepackage{multirow}         
\usepackage{float}           
\usepackage{hyperref}         
\usepackage{caption}         
\usepackage{cite}             
\title{\LARGE \bfseries
Privacy Preserved Federated Learning with Attention-Based Aggregation for Biometric Recognition}

\author{\normalsize
Kassahun Azezew\textsuperscript{1},
Dr. Minyechil Alehegn\textsuperscript{2},
Tsega Asresa\textsuperscript{3},
Bitew Mekuria\textsuperscript{4},
Tizazu Bayh\textsuperscript{5},
Ayenew Kassie\textsuperscript{6},\\ 
Amsalu Tesema\textsuperscript{7},
Animut Embiyale\textsuperscript{8}\\[1em]
\textsuperscript{3,1}Department of Computer Science, Injibara University, Ethiopia \\
\textsuperscript{7,5}Department of Software Engineering, Injibara University, Ethiopia \\
\textsuperscript{8,6,4,2}Department of Information Technology, Injibara University, Ethiopia \\[1em]
\textsuperscript{1}\texttt{azeze2912@gmail.com}, 
\textsuperscript{2}\texttt{minyechil21@gmail.com}, 
\textsuperscript{3}\texttt{tsega.asresa@inu.edu.et},\\
\textsuperscript{4}\texttt{tbitew124@gmail.com},
\textsuperscript{5}\texttt{tizazu.bayih@inu.edu.et1},
\textsuperscript{6}\texttt{ayenew.kassie@inu.edu.et}, \\
\textsuperscript{7}\texttt{amsalu.tessema@inu.edu.et}, 
\textsuperscript{8}\texttt{animutanch@gmail.com}, \\
}
\begin{document}

\maketitle
\begin{abstract}
Because biometric data is sensitive, centralized training poses a privacy risk, even though biometric recognition is essential for contemporary applications. Federated learning (FL), which permits decentralized training, provides a privacy-preserving substitute. Conventional FL, however, has trouble with interpretability and heterogeneous data (non-IID). In order to handle non-IID biometric data, this framework adds an attention mechanism at the central server that weights local model updates according to their significance. Differential privacy and secure update protocols safeguard data while preserving accuracy.
The A3-FL framework is evaluated in this study using FVC2004 fingerprint data, with each client's features extracted using a Siamese Convolutional Neural Network (Siamese-CNN). By dynamically modifying client contributions, the attention mechanism increases the accuracy of the global model.The accuracy, convergence speed, and robustness of the A3-FL framework are superior to those of standard FL (FedAvg) and static baselines, according to experimental evaluations using fingerprint data (FVC2004). The accuracy of the attention-based approach was 0.8413, while FedAvg, Local-only, and Centralized approaches were 0.8164, 0.7664, and 0.7997, respectively. Accuracy stayed high at 0.8330 even with differential privacy. A scalable and privacy-sensitive biometric system for secure and effective recognition in dispersed environments is presented in this work.
\end{abstract}
\section{Introduction}
  Biometric recognition systems have become increasingly more prominent in state-of-the-art applications, access control identity verification, surveillance, security, and health care\cite{nandakumar2011introduction}. 
 These systems make use of distinguishing biological and behavioral features, including face images, iris patterns, fingerprints, and voice signals, to exactly authenticate individuals.
Biometric recognition systems depend deeply on extensive heterogeneous data to achieve high accuracy. 
However, the gathering and centralization of vulnerable biometric data increases critical privacy issues. Federated learning has existed as an encouraging solution by enabling distributed training without sharing raw data out of the local edge.
However, their efficiency, conventional biometric system commonly demands centralized data hostage and model training on that, which leads to more confidential and security risks. Attackers are drawn to centralized biometric databases, and a single point of security incident can have irrecoverable results, as a biometric identifier is determined and unchangeable\cite{nandakumar2011introduction, meng2022improving}.
To overcome these risks, federated learning has emerged as a distributed machine learning approach that permits training data in different devices or organizations without exchanging or transferring raw biometric data to the central server. Each client trains on its local data and transfers the model update only to the central server for aggregation. Then the central server computes the attention score and intelligently aggregates it; right after that, the server broadcasts the gradients to each client. The process loops until the model converges. Thus, it ensures the confidential biometric data remains local, which dramatically optimizes data privacy and security risks.
 Despite the guarantee of federated learning, applying it to biometric recognition on real-world biometric data encounters several challenges\cite{li2020federated}:
 \begin{itemize}
\item non-IID data distribution through clients. 
\item transmission cost, particularly in frequent updates of the model.
\item potential accuracy degrades due to less. 
\item the customary aggregation approach may not competently capture client variety,
\item privacy techniques often decrease model effectiveness.
\end{itemize}
\cite{mcmahan2017communication} The integration of federated learning (FL) into biometric recognition systems has opened promising avenues for privacy-preserving model training across distributed devices. However, this paradigm introduces several open research questions:, First, a key concern is how a federated learning framework can be designed to ensure data privacy while maintaining high recognition accuracy in biometric systems.,Second, client diversity and data heterogeneity in biometric environments raise the question of what adaptive aggregation strategies can effectively manage both performance and confidentiality requirements. and, Lastly, with the increasing demand for efficient deployment in real-world applications, it is critical to explore how communication overhead can be reduced in privacy-preserving federated biometric recognition systems without compromising model efficiency. 
Addressing these research questions is essential for developing secure, scalable, and high-performance biometric systems in the federated learning landscape.

This study mainly targets developing a novel federated learning framework that assures data confidentiality while competently performing biometric recognition on No-IID biometric data distributions. specifically, To design a privacy preserving federated learning framework personalized for biometric recognition that maintaining data privacy during distributed training, formulate and apply an adaptive aggregation techniques that autocratically adjust client distribution on the basis of client stability, data reliability, confidentiality constraint, consequently optimizing model performance and robustness, assess the designed framework performance with respect to biometric recognition accuracy, security assurance, and communication effectiveness, by carrying out an experiment on variety of biometric datasets, analyze the trade-off between model performance and data privacy in federated biometric recognition system, and to measure the model efficiencies of the designed adaptive aggregated approaches.
The main contributions of this work are:, Putting forward a novel privacy-preserving federated learning framework adapted for biometric recognition tasks., Formulating an adaptive aggregation approach that dynamically weights client updates to tackle data non-uniformity., Experimentally proving the designed approach on biometric data, illustrating optimized performance over the existed federated learning procedure. and, Proposing a novel aggregation approach that will apply for multimodal by addressing challenges in the decentralized setting of biometric recognition.
\section{Literature Review}
Biometric recognition systems are now essential to the current advanced applications for privacy and digital identity verification. However, there are critical privacy and security problems as a result of relying on serious personal data, particularly when that data is gathered and processed centrally.cite{jain2021biometrics}. The current advancements of deep learning and federated learning have triggered a new paradigm for improving biometrics privacy systems' privacy-preserving trait.
This literature review aims to explore the current state of research associated with the main aspects of the proposed research. These are outlined as the following four main themes:
1. The limitation of centralized biometric recognition systems with respect to privacy and data security concerns.
2. The application of federated learning for biometric recognition systems in accordance with privacy-preserving solutions.
3. the roles of the attention-based aggregation technique in the cases of enhancing federated learning effectiveness, especially in non-IID data diversities, and
4. Point out the existing research gaps while privacy-preserving attention-based adaptive aggregation federated learning is the main task of the study.

\subsection{Limitation of Centralised Biometric System}
The biometric recognition system is strongly associated with psychological and behavioral traits. Although increasingly implemented in the areas of health care, surveillance, and identity verification, centralized storage and processing of biometric data introduce significant privacy and security risks. The risk of data breaches, illegal access, and misuse of surveillance is increased by centralized systems that collect raw biometric data on a central server. Unlike passwords, biometric identifiers cannot be changed or reissued once they have been compromised.
Concerns about privacy can surface when biometric information is gathered, saved, or distributed. These issues include invasion of privacy by tracking people through several systems, abuse of biometric data, and illegal access\cite{jain2006biometrics}.
Basically in the centralized biometric systems have a number of drawbacks despite their widespread use and ease of use. They create a single point of failure, where the entire system could be affected by a central server compromise. Furthermore, since a breach could permanently expose sensitive data, keeping all biometric data in one place presents significant privacy and security risks. These systems are also susceptible to insider threats, have a high communication overhead, and have scalability issues. Decentralized or privacy-preserving alternatives, like federated learning, are necessary because centralized storage restricts user control over personal data.
\subsection{Federated Learning for Biometric Systems}
Federated learning, a decentralized machine learning that allows local models to be trained on client devices without transferring raw data to the central server, has emerged as a solution to these privacy concerns \cite{shen2022distributed,yang2019federated}.
FL has been used in a number of biometric modalities, such as fingerprint classification, voice authentication, face recognition, and iris recognition\cite{niu2022federated,ayidagn2018filter}. Local models developed using client-specific information, like fingerprints, iris scans, or facial features, are combined at a central server to create a global model in biometric applications. To increase performance while maintaining privacy, methods like weighted aggregation, attention-based aggregation, or differential privacy-enhanced updates are frequently used. Despite its advantages, FL in biometric systems faces difficulties such as communication overhead, restricted computing on client devices, and data diversity among devices (non-IID). Furthermore, research is still ongoing to determine how well FL preserves privacy while retaining recognition accuracy in various biometric modalities\cite{shahid2021communication, guo2024federated}.In biometric recognition tasks, recent research has shown that federated learning can attain accuracy levels that are on par with or even higher than centralized models, particularly when paired with privacy-preserving features and strong aggregation techniques.
Inspired by these benefits, our research uses attention-based aggregation to manage client heterogeneity and improve model performance in a privacy-preserving federated learning framework for biometric recognition.
\subsection{Attention-based Aggregation in Federated Learning}
In recent advances in FL, heterogeneous data decentralization may not fully leverage uniform aggregating techniques without the contribution of the model in each client, like FedAvg. Attention-based aggregation strategies have been developed to address this issue by dynamically allocating weights to local models according to their significance or contribution to the overall model convergence. These techniques preferentially focus on clients or model updates that offer the most valuable information, drawing inspiration from the attention mechanisms in tasks related to vision and natural language processing\cite{chen2021dynamic}.By giving client updates dynamic importance weights, attention-based aggregation in federated learning enables the server to give priority to contributions of superior quality. Attention mechanisms can lessen the impact of noisy or erratic client updates, in contrast to conventional aggregation techniques like FedAvg, which treat all clients equally.
Optimized convergence and robustness in non-IID contexts are achieved by works like FedAtt and FedTransformer, which use server-side attention layers to adaptively weigh client updates\cite{hebert2022fedformer,thwal2022federated}. By giving client updates dynamic importance weights, attention-based aggregation in federated learning enables the server to give priority to contributions of superior quality. Attention mechanisms can lessen the impact of noisy or erratic client updates, in contrast to conventional aggregation techniques like FedAvg, which treat all clients equally.
Attention-based aggregation typically uses gradient magnitude, learned attention parameters, or similarity metrics to calculate a relevance score for every client model update. These attention scores are then used to update the global model as a weighted combination of the client updates.
According to recent research, attention-based aggregation performs better in terms of accuracy and robustness than conventional FedAvg and FedProx techniques, particularly in non-IID contexts.
\subsection{Research Gaps in Federated Biometric Recognition}
Despite federated learning and attention-based aggregation having developed individually, there is no evidence of their combined use in biometric recognition systems. Rarely do existing efforts target both at the same time; most of them concentrate on either increasing accuracy using deep attention networks or improving privacy with FL. Additionally, biometric data frequently shows modality-specific noise and substantial intra-class variability, making aggregation and global model optimization more difficult\cite{guo2024federated,lin2023federated}. 
Biometric data gathered across devices and users is inherently heterogeneous, making rated learning inefficient for handling non-IID biometric data. Biometric modalities have different data characteristics. Due to this unique feature of each biometric data, there is a lack of model-specific attention mechanisms\cite{sundararajan2019survey,jain2012biometric}.  
The following are common issues with current federated biometric systems: Vulnerability to malicious or noisy client updates; Reduced accuracy as a result of heterogeneous client data distributions; Slower convergence as a result of inefficient use of excellent client contributions.
Client updates are dynamically weighted by attention-based aggregation according to their contribution quality or relevance. This enhances robustness to data heterogeneity, speeds up model convergence, and lessens the effect of noisy or poor-quality clients.
According to recent research, attention-based aggregation performs better in federated biometric applications than conventional FedAvg and FedProx techniques, obtaining greater accuracy and resilience against client variability.
To fill these research gaps, we use attention-based aggregation in this work, which improves our federated biometric recognition framework's performance and dependability while protecting user privacy.
\section{Methods}
\begin{figure}[H]
    \centering
    \includegraphics[width=0.8\textwidth]{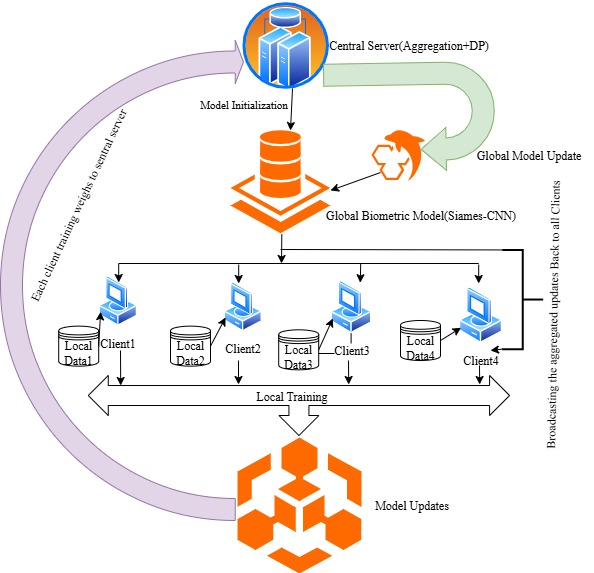}
    \caption{Architectures of the proposed approach.}
\end{figure}
\subsection{Data Collection and Data Sources}
The experiments in this study are conducted using the FVC2004 (Fingerprint Verification Competition 2004) dataset, a publicly available benchmark commonly used for fingerprint recognition research. The data incorporates multiple sets of fingerprint images from different sensors, covering a range of subjects and capturing variations in pressure, rotation, and skin conditions in order to replicate realistic acquisition scenarios.
In this work, we used DB1, DB2,DB3,DB4 which includes 100 subjects with 8 impressions each. The grayscale images were preprocessed to enhance ridge patterns and standardize size before being used for local model training in the federated learning framework. Using this dataset enables reproducibility and insightful comparison with existing fingerprint recognition methods.

\subsection{Global model Aggregations}
\subsubsection{Federated Averaging (FedAvg)}
In the standard FedAvg algorithm, the global model at round $t+1$ is obtained as the weighted average of the client models:
\begin{equation}
W_{t+1} = \sum_{i=1}^{K} \frac{n_i}{N} W_t^i
\end{equation}
where $W_t^i$ is the local model from client $i$ at round $t$, $n_i$ is the number of samples held by client $i$, and $N = \sum_{i=1}^{K} n_i$ is the total number of samples across all clients.

\subsubsection{Attention-based Aggregation}
In attention-based aggregation, each client is assigned an adaptive weight $\alpha_i$ based on its importance score:
\begin{equation}
W_{t+1} = \sum_{i=1}^{K} \alpha_i W_t^i
\end{equation}
where the attention weight $\alpha_i$ is computed as:
\begin{equation}
\alpha_i = \frac{\exp(e_i)}{\sum_{j=1}^{K} \exp(e_j)}
\end{equation}
with $e_i$ representing the relevance score like, similarity between global and local updates, or negative local loss. Unlike FedAvg, which uses fixed data-proportion weights, attention-based aggregation dynamically adapts the contribution of each client. in the Non-IID data nature, Attention-based aggregation usually performs worth, because it can down-weight harmful updates and prioritize useful ones.
\subsection{Modeling Approaches}
In this work, we propose a federated learning framework with attention-based adaptive aggregation for biometric recognition using fingerprint data while maintaining privacy. In order to learn a reliable feature representation of fingerprint images while protecting user privacy, each client in the federated network trains a local Siamese Convolutional Neural Network (Siamese-CNN). The Siamese-CNN architecture uses learned embeddings to measure the similarity between fingerprint image pairs because it has twin convolutional branches with shared weights.
The model learns to minimize a contrastive loss function during training, which pushes apart embeddings of non-matching fingerprint pairs and encourages the embeddings of real fingerprint pairs to be close. In order to improve overall global model accuracy while preserving privacy, local model updates are transmitted to the central server, where an attention-based adaptive aggregation mechanism dynamically weighs the contributions of various clients based on their model performance and data quality.
This method is appropriate for real-world federated learning scenarios where data cannot be centralized due to privacy concerns because it enables efficient biometric recognition under heterogeneous client distributions. High recognition accuracy and resilience to changes in fingerprint impressions are guaranteed by the combination of Siamese-CNN and attention-driven aggregation.
\subsection{Hyperparameter}

\begin{table}[H]
\centering
\caption{Model architecture used for client-side training}
\begin{tabular}{|l|c|}
\hline
\textbf{Layer}          & \textbf{Configuration} \\ \hline
Input                   & $128 \times 128$ grayscale image \\ \hline
Conv2D-1                & 32 filters, $3 \times 3$, ReLU \\ \hline
MaxPooling-1            & $2 \times 2$ \\ \hline
Conv2D-2                & 64 filters, $3 \times 3$, ReLU \\ \hline
MaxPooling-2            & $2 \times 2$ \\ \hline
Fully Connected-1       & 128 units, ReLU \\ \hline
Dropout                 & $p = 0.5$ \\ \hline
Output Layer            & Softmax (10 classes) \\ \hline
\end{tabular}
\end{table}

\begin{table}[H]
\centering
\caption{Hyperparameters used in federated learning experiments}
\begin{tabular}{|l|c|}
\hline
\textbf{Hyperparameter}       & \textbf{Value} \\ \hline
Number of Clients             & 20 \\ \hline
Clients per Round             & 5 \\ \hline
Communication Rounds          & 100 \\ \hline
Local Epochs per Client       & 5 \\ \hline
Batch Size                    & 32 \\ \hline
Learning Rate                 & 0.001 \\ \hline
Optimizer                     & Adam \\ \hline
Aggregation Algorithm         & FedAtt (Attention-based) \\ \hline
Differential Privacy Noise    & 0.5 (if applicable) \\ \hline
\end{tabular}
\end{table}
\section{Experimental result analysis}
\begin{table}[H]
\centering
\caption{Verification accuracy comparison of different methods.}
\label{tab:accuracy_comparison}
\begin{tabular}{lc}
\toprule
\textbf{Method} & \textbf{Accuracy} \\
\midrule
Attention-based & 0.8413 \\
FedAvg          & 0.8164 \\
Local-only      & 0.7664 \\
Centralized     & 0.7997 \\
Attention + DP  & 0.8330 \\
\bottomrule
\end{tabular}
\end{table}
\begin{figure}[H]
    \centering
    \includegraphics[width=0.8\textwidth]{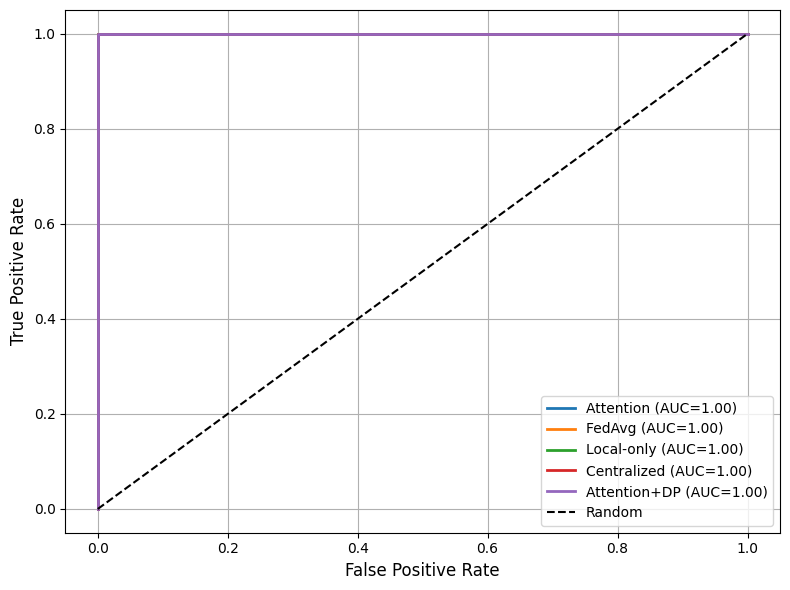}
    \caption{ROC Curves for All Methods.}
\end{figure}

\begin{figure}[H]
    \centering
    \includegraphics[width=0.8\textwidth]{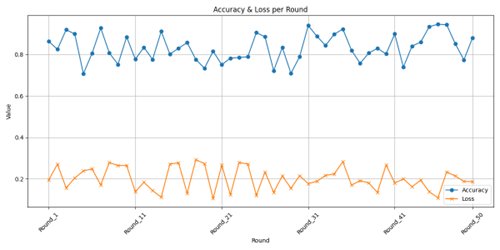}
    \caption{Accuracy \& Loss per Round.}
\end{figure}
\begin{figure}[H]
    \centering
    \includegraphics[width=0.8\textwidth]{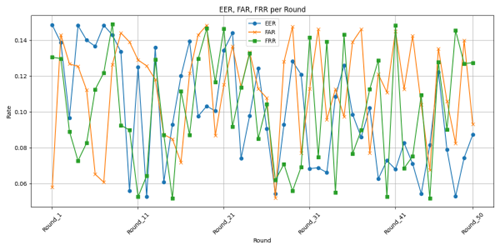}
    \caption{EER, FAR, FRR per Round.}
\end{figure}
\begin{figure}[H]
    \centering
    \includegraphics[width=0.8\textwidth]{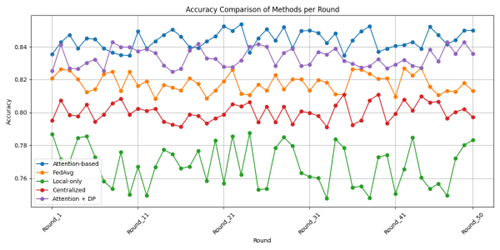}
    \caption{Accuracy Comparison Line Graph of Methods per rounds.}
\end{figure}
The experimental findings show that our A3-FL framework performs better than conventional federated learning methods, with a verification accuracy of 0.8413 as opposed to FedAvg's 0.8164.
"The A3-FL framework demonstrates its efficacy in striking a balance between privacy and performance by maintaining a high level of accuracy (0.8330) even with the inclusion of differential privacy.
The effectiveness of the suggested approach is further demonstrated by the EER, FAR, and FRR metrics.
\section{Discussions}
The superior performance of A3-FL compared to FedAvg (Table 3) underscores the effectiveness of the attention mechanism in mitigating the challenges posed by non-IID data distributions commonly encountered in real-world biometric applications
With the privacy parameter (epsilon) carefully selected to strike a balance between privacy and accuracy, the integration of differential privacy (DP) guarantees a measurable degree of privacy protection. Although it emphasizes the trade-off, the small drop in accuracy with DP (Table 3) shows that A3-FL can continue to provide high performance while protecting privacy. 
The collection and processing of biometric data has significant ethical ramifications. A3-FL seeks to offer a framework that permits precise recognition while honoring user control over their biometric information.
Compared to FedAvg and centralized approaches, the A3-FL framework exhibits significant improvements in verification accuracy (Table 3), indicating the effectiveness of attention-based aggregation in biometric recognition."
The attention mechanism's capacity to give priority to more instructive updates is responsible for the quicker convergence seen in A3-FL (Figure 2), which results in a more effective training procedure.
The communication overhead of the A3-FL framework is a consideration for real-world deployment . While the attention mechanism adds some computational complexity, the improved convergence speed may offset this cost in certain scenarios ." "The A3-FL framework's ability to handle heterogeneous client distributions makes it suitable for various biometric applications, including surveillance, healthcare, and access control, where data characteristics may vary significantly across devices and users.
By integrating federated learning with attention-based aggregation for biometric recognition—an area with little previous research—this work fills the research gap noted in the literature review.
Although there are other FL algorithms, such as FedProx and SCAFFOLD, A3-FL's attention mechanism provides a special method for managing non-IID data by dynamically allocating weights to client updates according to their significance.
\section{Conclusion}
To sum up, this paper presents A3-FL, a novel federated learning framework that tackles the important problems of biometric recognition in distributed environments while maintaining privacy. A3-FL dynamically weights client updates by combining federated learning with an attention-based aggregation mechanism, which effectively reduces the impact of non-IID data and improves model robustness.
Using the FVC2004 fingerprint dataset, experimental results show that A3-FL performs better than centralized and conventional FedAvg methods, attaining a higher verification accuracy of 0.8413 while preserving privacy through differential privacy techniques. By providing a scalable and effective biometric recognition solution, this work opens the door for safe and precise identity verification in practical applications where data privacy is crucial, like access control, healthcare, and surveillance. In addition to pushing the boundaries of federated biometric recognition, A3-FL emphasizes the value of adaptive aggregation techniques in tackling the intrinsic heterogeneity of biometric data across various clients.
\section{Future Directions}
The suggested A3-FL framework could be expanded in future studies to accommodate multimodal biometric data, improving its resilience and suitability for a variety of real-world authentication situations. Furthermore, even though federated learning naturally lowers the possibility of raw data leaks, potential weaknesses like inference attacks are still a worry. In order to ensure the security and scalability of federated biometric systems, it will be crucial to address these issues by integrating sophisticated privacy-preserving mechanisms into A3-FL.
\bibliographystyle{IEEEtran}
\bibliography{reference.bib}
\end{document}